%% file: egpaper.tex
\documentclass[10pt,twocolumn,letterpaper]{article}

\usepackage{wacv}
\usepackage{times}
\usepackage{epsfig}
\usepackage{graphicx}
\usepackage{amsmath}
\usepackage{amssymb}
\usepackage{booktabs}
\usepackage{xcolor}
\usepackage{microtype}
\usepackage{makecell}
\usepackage{caption} 
\def\wacvPaperID{226} 

\wacvfinalcopy 

\ifwacvfinal
\def\assignedStartPage{1} 
\fi

\ifwacvfinal
\usepackage[breaklinks=true,bookmarks=false]{hyperref}
\else
\usepackage[pagebackref=true,breaklinks=true,colorlinks,bookmarks=false]{hyperref}
\fi

\ifwacvfinal
\else
\fi

\begin{document}

\title{Billion-Scale Pretraining with Vision Transformers for \\ Multi-Task Visual Representations}

\author{Josh Beal \qquad Hao-Yu Wu \qquad Dong Huk Park \qquad Andrew Zhai \qquad Dmitry Kislyuk\\
Pinterest\\
{\tt\small \{jbeal, rexwu, dhukpark, andrew, dkislyuk\}@pinterest.com}
}

\maketitle

\begin{abstract}
Large-scale pretraining of visual representations has led to state-of-the-art performance on a range of benchmark computer vision tasks, yet the benefits of these techniques at extreme scale in complex production systems has been relatively unexplored. 
We consider the case of a popular visual discovery product, where these representations are trained with multi-task learning, from use-case specific visual understanding (e.g. skin tone classification) to general representation learning for all visual content (e.g. embeddings for retrieval). 
In this work, we describe how we (1) generate a dataset with over a billion images via large weakly-supervised pretraining to improve the performance of these visual representations, and (2) leverage Transformers to replace the traditional convolutional backbone, with insights into both system and performance improvements, especially at 1B+ image scale. 
To support this backbone model, we detail a systematic approach to deriving weakly-supervised image annotations from heterogenous text signals, demonstrating the benefits of clustering techniques to handle the long-tail distribution of image labels.
Through a comprehensive study of offline and online evaluation, we show that large-scale Transformer-based pretraining provides significant benefits to industry computer vision applications. The model is deployed in a production visual shopping system, with 36\% improvement in top-1 relevance and 23\% improvement in click-through volume. We conduct extensive experiments to better understand the empirical relationships between Transformer-based architectures, dataset scale, and the performance of production vision systems.
\end{abstract}

\section{Introduction}
Visual representation learning is a core foundation in online content search and recommendation systems~\cite{Zhai2019-qj, Bell2020-cg, relatedpins}. In recent years, we have increasingly seen the paradigm of training a single, high-capacity, deep neural network model jointly across many heterogeneous tasks, especially in industry settings with complex use-cases~\cite{text2text, Zhai2019-qj, Bell2020-cg}. There are multiple benefits, including (1) the ability to jointly utilize large amounts of strongly and weakly supervised data, instead of training domain-specific models in isolation, sometimes leading to better performance compared with domain-specific models~\cite{Zhai2019-qj}, (2) an infrastructure benefit due to reduced operational overhead to maintain fewer models, and (3) a reliable source model to build upon for downstream modeling tasks via transfer learning. 

Many search, recommendation, and content understanding tasks require a representation capturing both visual and semantic components (Figure \ref{fig:modeloverview}). This work focuses on the single multi-task image representation model powering visual understanding for a widely-used visual discovery product, referred to as the ``Unified Visual Embedding''. This model powers tens of production use cases and is an area of substantial investment. Some production use-cases include: 
\begin{itemize}
    \item \textbf{Retrieval:} the embeddings are used through Approximate Nearest Neighbors in several retrieval systems, including Visual Search and Visual Shopping.
    \item \textbf{Features:}
    the embeddings are used as features in other content understanding models that need information extracted from images. Such models can be purely visual (e.g. skin-tone classification), aggregating multiple images (e.q. video understanding), or combining other signals such as text and graph structure~\cite{Zhuang2019-kk, ying2018graph, yang2020multisage}. It is a computationally efficient way to leverage visual signals in model training and serving.
\end{itemize}

In our paper we describe the implementation, experimentation, and productionization of large scale pretraining on over a billion images, along with the adoption and deployment of a Transformer-based architecture (Vision Transformers), supplanting a CNN architecture approach that has been an industry standard in recent years.
We focus on a weakly supervised label preparation methodology along with ablations including dataset size variation to construct the critical dataset \textbf{Annotations-1.3B}, and describe how our billion-scale pretrained Vision Transformer model benefits the multi-task image representation model, along with the end-to-end production system relevance, and engagement impact through human judgement and A/B experimentation.
We conclude with insights into generalization (few-shot, cross-domain) performance. To the best of our knowledge, this is the first large-scale industry application of pure Transformer-based models for image retrieval.

\section{Related Work}

\subsection{Visual Search Systems}

Visual search has been widely adopted in social and e-commerce applications, including Facebook \cite{Tang2019-we}, Pinterest \cite{Zhai2019-qj}, eBay \cite{10.1145/3097983.3098162}, Google, Microsoft \cite{Hu2018-yq}, Alibaba \cite{Zhao2019-im, Zhou2020-hg}, and Amazon \cite{Zuo2020-kh}. In prior work, GrokNet \cite{Bell2020-cg} and Shop the Look \cite{Shiau2020-lc} have described the recent details of industrial-scale visual search systems. These visual search systems leverage multi-task learning to optimize a single embedding for different applications in the organization. The visual embedding models powering these systems are trained using deep metric learning, including classification-based methods \cite{Movshovitz-Attias2017-ig, Zhai2018-kr} and triplet-loss-based methods \cite{Hoffer2015-ez, Wu2017-oe}.

\subsection{Large-Scale Pretraining}

Large-scale image datasets have proven to be useful in the pretraining of general visual representations. Some of the earliest work in this area involved pretraining AlexNet~\cite{krizhevsky2012imagenet} on YFCC100M~\cite{10.1145/2812802}, showing that pretraining on a large dataset of Flickr images could perform comparably to ILSVRC-2012~\cite{deng2009imagenet}, a fully supervised dataset of 1.3 million images. The quality of the text metadata has limited the effectiveness of YFCC100M, and in subsequent work, many of the improvements stem from the selection of higher quality sources of textual supervision. JFT-300M~\cite{sun2017revisiting} is a large-scale dataset developed at Google, consisting of around 300M images and 18k labels in the dataset taxonomy. The labels for each image are derived from a mixture of web signals. Recently this dataset has been leveraged in the BigTransfer~\cite{kolesnikov2019big} work to achieve state-of-the-art results on ILSVRC-2012 (hereafter referred to as ImageNet-1k) and the Visual Task Adaptation Benchmark (VTAB)~\cite{Zhai2019-vd}. Facebook \cite{mahajan2018exploring} investigated pretraining on billions of Instagram images using hashtags as a form of weak supervision, achieving competitive ImageNet-1k performance. Large subsets of Instagram images, selected randomly without bias to the hashtag distribution, have also been useful in the context of self-supervised learning \cite{Caron2020-zg}. OpenAI has explored the effectiveness of image-to-caption matching in CLIP~\cite{Radford_2021}, introducing the WebImageText dataset of approximately 400 million (image, text) pairs. The webly-supervised pretraining paradigm offers benefits to more complex vision tasks, such as object detection \cite{beal2020toward} and video understanding \cite{Stroud2020-yj}. Similar to CLIP, the task of pairing videos with their associated title, description, and other metadata, has led to state-of-the-art results on action recognition tasks when using a dataset of 70 million YouTube videos \cite{carreira2019short}.

\begin{figure*}
\begin{center}
\includegraphics[width=\textwidth]{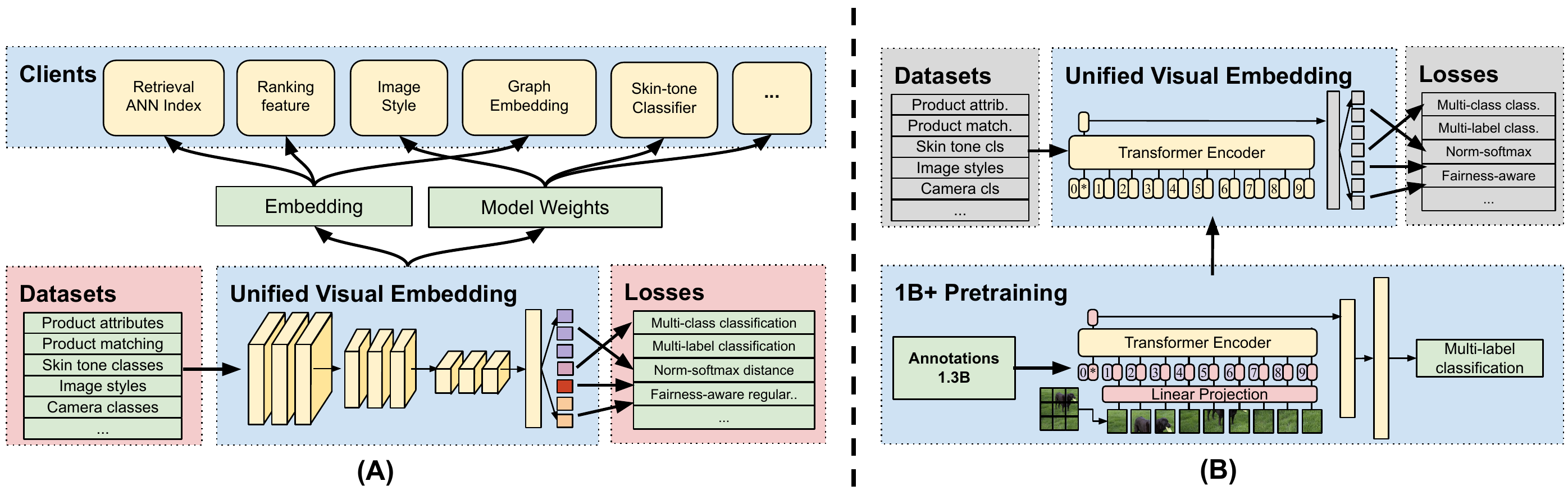}
  \caption{(A) The overall architecture for Unified Visual Embeddings~\cite{Zhai2019-qj}, consisting of one backbone convolutional neural network model consuming a variety of datasets including classification and metric learning across a set of loss and regularization functions. The embedding is consumed by a variety of customers across retrieval, as an input feature, and for fine-tuning domain-specific models. We also show (B) our proposed methodology with grey boxes denoting unchanged components. We introduce billion-scale image pretraining to produce a strong backbone encoder and leverage the Vision Transformer encoder as a replacement of the CNN backbone. }
  \label{fig:modeloverview}
\end{center}
\end{figure*}

\subsection{Transformer Architectures}
Transformer~\cite{vaswani2017attention} architectures have become the state-of-the-art solution for many natural language processing tasks, based on a simple and scalable application of multi-head attention. BERT~\cite{devlin2018bert} and GPT-3~\cite{brown2020language} have demonstrated that these architectures have a large pretraining capacity, with few signs of saturating performance as the dataset size and model size continue to increase. Fully Transformer-based architectures have recently been successful for image classification, with Vision Transformers~\cite{dosovitskiy_arxiv2020, touvron2020training} achieving state-of-the-art ImageNet-1k performance, while offering compelling advantages in training compute and memory efficiency relative to the equivalent ResNet~\cite{he2016deep} architectures.

\section{Methodology}
\subsection{System Overview}
Figure~\ref{fig:modeloverview} (A) describes the overall setup for Unified Visual Embeddings, which includes optimization objectives spanning across many modes of classification (single-class, multi-label, multi-label softmax~\cite{mahajan2018exploring}), metric learning (normalized sampled softmax~\cite{Zhai2018-kr}, distance weighted sampling~\cite{Wu2017-oe}) and auxiliary regularization losses.
To evaluate model improvements, we compare both qualitative and quantitative evaluation metrics across tens of datasets. Visual Search~\cite{jing15} remains one of the most important applications of unified visual embeddings, where performance is predominantly determined by the representation quality and is the focus of our offline evaluations and A/B experiments.

Figure~\ref{fig:modeloverview} (B) describes an overview of our modifications to the prior unified embedding setup with our billion-scale weakly supervised pretraining and end-to-end Transformer encoder. We go into details of our methodology below, describing these two components that each led to significant improvements across the majority of the evaluation metrics.

\subsection{Billion-Scale Weakly Supervised Dataset}
Motivated by prior work leveraging user-provided text for large-scale weak supervision~\cite{mahajan2018exploring}, we leverage multiple text understanding models to create our billion-scale dataset. We apply term clustering techniques and filter the candidates according to visual concreteness and top-level interest-match in order to create our \textbf{Annotations-1.3B} dataset.

\subsubsection{Label Generation}
\label{sec:label_generation}

\begin{figure}
  \centering
  \includegraphics[width=0.95\linewidth]{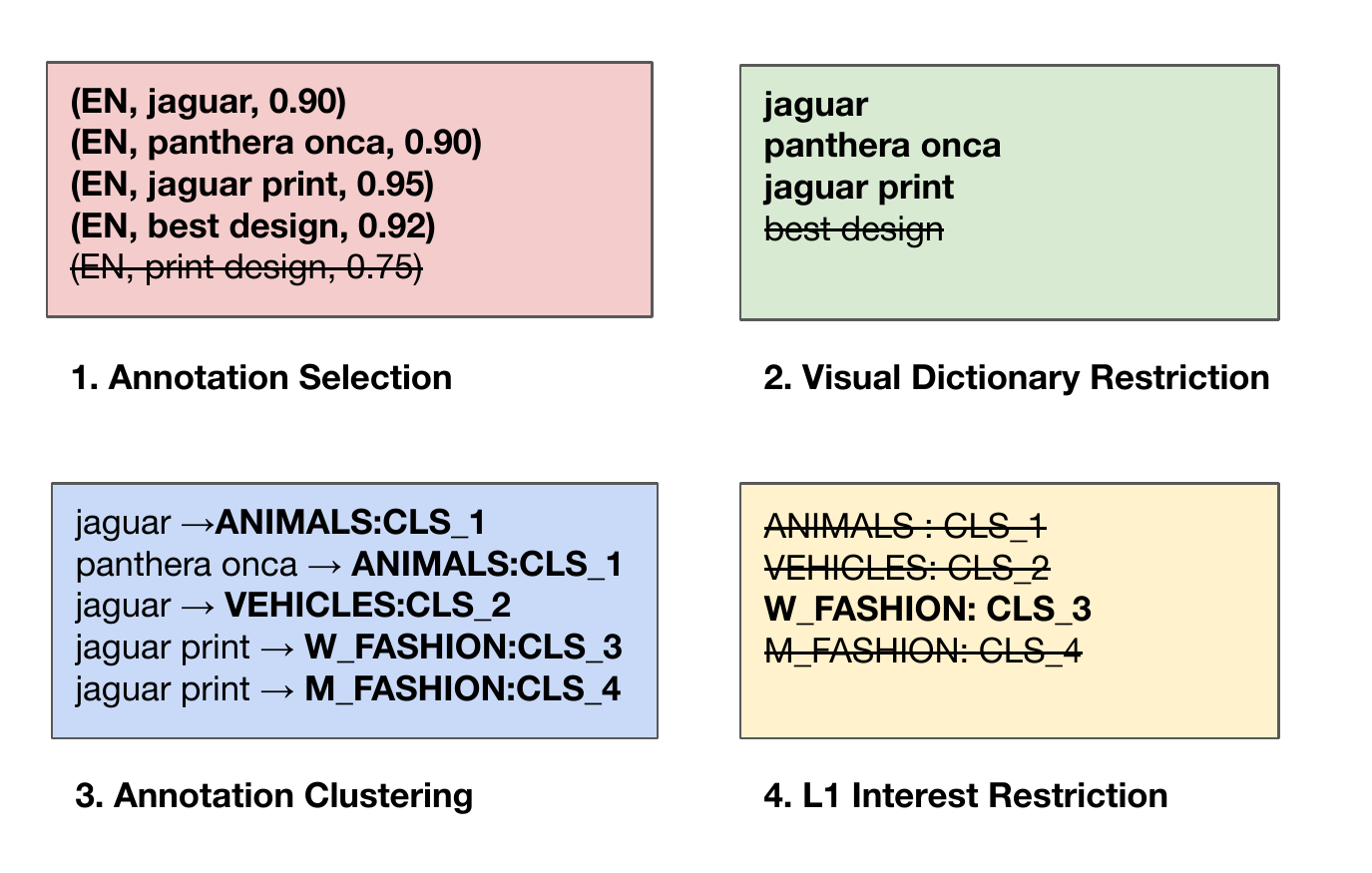}
  \vspace{-0.2cm}
  \caption{Overview of the label generation procedure.}
  \label{fig:labelgen}
\end{figure}

We leverage multiple content understanding models to derive weakly-supervised image annotations from over a billion web images. Figure~\ref{fig:labelgen} depicts a high-level overview of our process, which is described in detail below.

\paragraph{Annotation Selection:} We use  high-confidence outputs of a keyword prediction model as the basis for our label generation method. To simplify the subsequent steps in the pipeline, we select only the canonical, non-sensitive English keywords, referred to as ``annotations.'' Furthermore, we select a confidence score threshold such that annotations are discarded if their score falls below the threshold. This ensures a high level of precision in the resulting output. Similar to hashtag-based pretraining approaches~\cite{mahajan2018exploring}, it is possible that applicable annotations are missing from the signal output due to incompleteness of the text candidate sources (e.g., content is missing a title and description).

\paragraph{Visual Dictionary Restriction:} One challenge in working with the predicted annotations is that some terms may be helpful for general content understanding, but are less helpful for learning visual features due to the abstract nature of the terms. The relationship between the visual concreteness of terms and the ability of machine learning algorithms to learn cross-modal relationships based on these terms has been investigated in prior work~\cite{mahajan2018exploring, DBLP:conf/naacl/HesselML18}. However, manually assigning a concreteness score to each term in the annotations dictionary would involve a large amount of human effort. Therefore, we developed an algorithmic approach to defining a concreteness score for each term.

For each term in the annotations dictionary, we train a semi-supervised binary classifier to distinguish the applicability of that term to a given image. We sample images with an annotation matching that term at score greater than the high-confidence threshold, and sample an equal number of random negative examples that do not have a matching annotation. We train and evaluate a lightweight embedding-based MLP classifier for each term using the production unified visual embedding as input to the model. The top-1 accuracy of the term's classifier serves as the visual concreteness score for the term, since the performance reflects the general ability of a model to distinguish the applicability of the term based on the visual information only.

\begin{figure}
  \centering
  \includegraphics[width=0.9\linewidth]{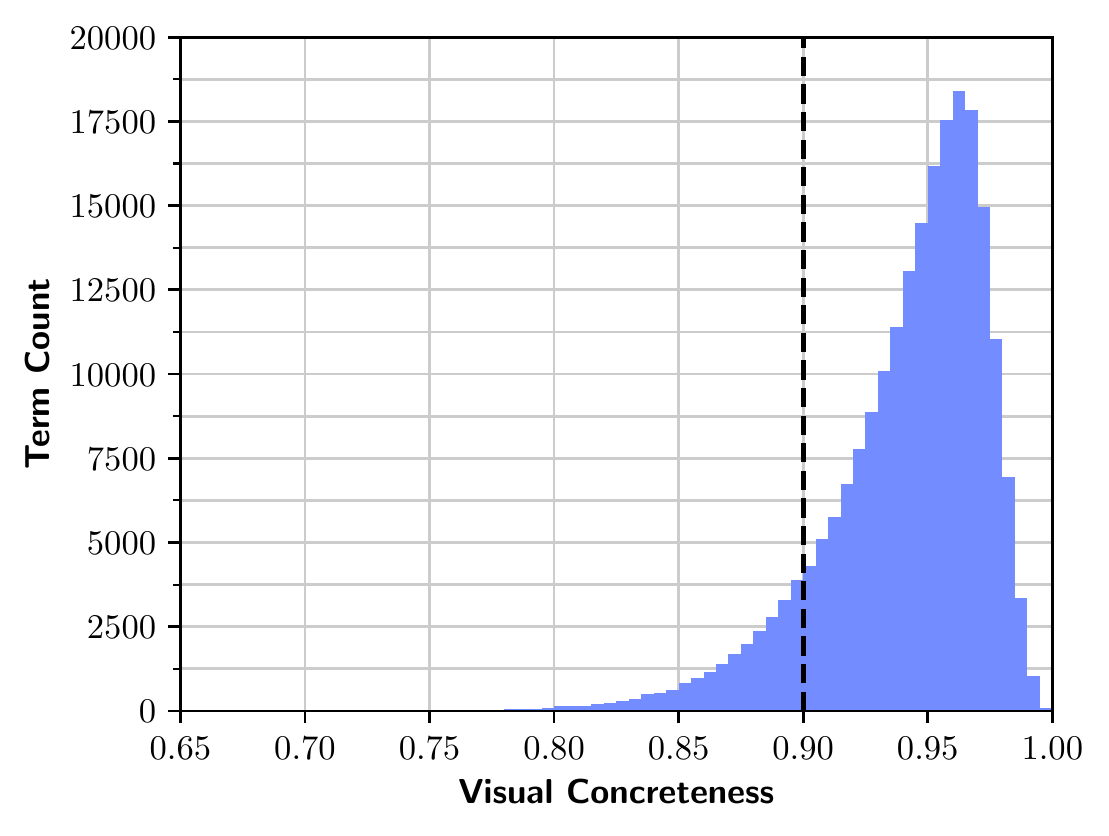}
  \caption{Visual concreteness distribution for terms in the annotations dictionary and visual dictionary threshold.}
  \label{fig:visual_concreteness}
\end{figure}

Applying this method, we analyze the visual concreteness of 218,879 terms in the annotations dictionary. In Figure~\ref{fig:visual_concreteness}, we present the distribution of the concreteness scores. We identify a decision boundary for the concreteness scores based on a quantitative analysis of the pretraining performance after removal of the low-quality terms from the dictionary. The visual dictionary consists of all terms not falling below the decision boundary. 
During the label generation procedure, if an annotation is not present in the visual dictionary, it is removed from further processing.

\paragraph{Annotation Clustering:} Another challenge with the predicted annotations is the large size of the label space, which consists of hundreds of thousands of terms. This creates practical challenges for training and motivates the use of clustering techniques to reduce the size of the label space. Our method first uses a classification model to map each annotation to its corresponding top-level interests (``L1 interests''). There are 24 high level interests (e.g., home decor, food and drinks, fashion) in the human-curated taxonomy. Within each of the L1 interest groups, k-means clustering is applied to the text embeddings of the terms within that interest group in order to yield a set of cluster identifiers. Each image is mapped to the cluster identifiers corresponding to its previously selected annotations. Note that each annotation may map to multiple cluster identifiers, since each annotation may map to multiple L1 interests.

\paragraph{L1 Interest Restriction:} We apply a classification model to map each image to top-level interests (``L1 interests``) based on the image's associated text metadata and user interactions. We remove the cluster identifiers where the L1 interest of that cluster fails to match the L1 interest of the image. This helps to address polysemy in the annotations space. Per the example in Figure \ref{fig:labelgen}, the keyword ``jaguar'' could be used in reference to an animal, a vehicle, or a fashion item, whereas the top-level interests for the image help to reduce the ambiguity of the textual content.

\subsection{Training Datasets}
We follow the label generation procedure to derive the Annotations-1.3B pretraining dataset from the application's image corpus. This dataset consists of 1329M images with 18k labels in the taxonomy and approximately 2.88 labels per image. We use a near-duplicate removal process to ensure that each image that is present in the training dataset is unique and non-overlapping with the validation sets.

The pretrained backbone network is used for initialization of the encoder in the Unified Embedding fine-tuning process. The multi-task training setup for the Unified Embedding follows the previous description in \cite{Zhai2019-qj, Shiau2020-lc}. Compared to the prior multi-task setting \cite{Shiau2020-lc}, there are three additional classification datasets: Image Style, Skin Tone, and Home Decor Color. Figure \ref{fig:modeloverview} depicts the relationship between the pretraining setting and the multi-task training setting.

\subsection{Model Architecture}
We experiment with two families of model architectures: CNN-based ResNext architectures~\cite{xie2017aggregated} and Transformer-based ViT architecture~\cite{dosovitskiy_arxiv2020, touvron2020training}. In this work, we use the ResNeXt-101 32x8d variant for all our CNN-based experiments (as it is the baseline in production), thus we will refer to it in short as ResNeXt-101 throughout this paper. We use the ViT-Base model variants in our experiments, which contain 12 Transformer encoder layers, 12 attention heads, hidden size of 768, and MLP size of 3072 (86M parameters)~\cite{dosovitskiy_arxiv2020}.
We consider the ViT-B/32 and ViT-B/16 variants, which use an image patch size of 32x32 and 16x16, respectively. Due to limitations on GPU memory and compute, and the quadratic complexity of the Vision Transformer model with respect to input sequence length, it was not practical to evaluate variants with smaller patch sizes. For pretraining, we use a prediction head trained with the multi-label softmax loss~\cite{mahajan2018exploring}, which yielded better results than training with per-cluster sigmoid outputs and binary cross entropy loss.

\section{Experiments}

\subsection{Offline Evaluation}
\input{tables/ue-pretraining-ablation}
\input{tables/label-generation-ablation}
\input{tables/ue-resolution-ablation}

We studied the effect of large-scale pretraining through the transfer performance of the fine-tuned Unified Visual Embedding on the key downstream tasks.

\subsubsection{Retrieval Evaluation}
We use three offline evaluation sets described in the previous works \cite{Zhai2019-qj, Shiau2020-lc} to measure the image retrieval performance of the fine-tuned Unified Visual Embedding: Visual Shopping (VS), Flashlight (F), and Lens (L). The metric average Precision@20 is used for Lens and Flashlight tasks, and metric P@1 is used for fine-grained Visual Shopping task. In this work, we use the methodology described in Shiau et al. \cite{Shiau2020-lc} to collect a Visual Shopping offline evaluation set with $\sim$40k (query, product) pairs plus $\sim$110k distractor set.

In Table~\ref{tab:ue_pretraining_ablation}, we see that improvements to the pretraining procedure yield consistent benefits for the CNN-based backbone. Switching from ImageNet-1k to a generic large-scale dataset (IG-940M) yielded an average relative improvement of +15.5\% across the three retrieval tasks of interest. When using a domain-specific large-scale dataset of comparable scale (Annotations-1.3B), we obtained a further +8.5\% average relative improvement to the retrieval performance. Finally, by leveraging the Vision Transformer model architecture with large pretraining capacity, we obtained an additional +6.7\% average relative improvement to the retrieval performance. The finding is consistent with ~\cite{dosovitskiy_arxiv2020, touvron2020training}, when the pretraining dataset is small, CNN-based backbone has slight advantage. Vision Transformer starts to outperform when we use the  Annotation-1.3B pretraining dataset. 

We studied the impact of design choices in the label generation procedure in Table~\ref{tab:label_generation_ablations}. The decisions to limit the selected terms to those in the visual dictionary and restrict the selected clusters to those matching the top-level interests of the content item yielded improvements to the Visual Shopping retrieval performance. Furthermore, reducing the size of the clustered label space benefits the computational efficiency of the prediction head in pretraining.

Per the results in Table~\ref{tab:ue_resolution_ablation}, we found that the Vision Transformer backbone can take advantage of higher-resolution inputs (384x384 images) in the fine-tuning process, whereas the ResNeXt-101 backbone did not see consistent benefits from higher-resolution inputs. 

\subsubsection{Classification Evaluation}
We observed consistent improvements across classification tasks when replacing the ImageNet-1k backbone with one that is pretrained on a large-scale dataset. For this evaluation, we measured the average P@1 across 10 tasks, namely, Camera Categories, Home Decor Color, Fashion Color, Pattern, Fabric, Image Style, Lowerbody Length, Dress Style, and Skin Tone. We found that using a ResNeXt-101 model pretrained on a generic large-scale dataset (IG-940M) yielded an average improvement of +2.7\% across these tasks, while pretraining on a domain-specific large-scale dataset (Annotations-1.3B) yielded a larger average improvement of +3.6\% over the ImageNet-1k baseline model. By adopting the ViT-B/16 backbone, we obtained the largest average improvement of +4.5\% over the baseline model. 

Furthermore, we observed large improvements for important downstream tasks where no training data had yet been integrated. The unified visual embedding is a key part of a near-duplicate image detection system~\cite{gusev2020evolution}, and we measure the performance of the visual embedding for this application using R@P95 on a held-out eval set. For the ResNeXt-101 model architecture, performance improved from 42.4\% to 60.8\% as the pretraining dataset is varied from ImageNet-1k to IG-940M. Furthermore, we found that as the pretraining dataset is varied from IG-940M to Annotations-1.3B, performance improved again from 60.8\% to 84.9\%. This task improvement was achieved without any specific data collection or optimization for the target use case.

\subsection{Human Relevance}
\input{tables/ue-e2e-relevance}
We deployed large-scale pretrained models to a visual shopping system, and measured the real-world relevance improvement with different pretraining datasets and model architectures through end-to-end human relevance evaluation as shown in Table ~\ref{tab:ue_e2e_relevance_eval}. We sampled roughly 8,000 traffic weighted user image queries in production, with half of the queries from fashion and half from home decor domain. For each model variant, we compute binary embeddings for the queries and retrieve top-1 results from the shopping product corpus. The pairs of (query, product) are then rated by in-house human evaluators. We reported two levels of relevance. The pair is rated \textit{extremely similar} when all key attributes, such as color, pattern and materials, match. The pair is rated \textit{similar} when 1-2 attributes have minor mismatches.

The general improvement trend found in offline evaluation experiments holds true for the industry scale visual shopping application. With CNN-based backbone, when we update the pretraining dataset from ImageNet-1k, to IG-940M, to Annotations-1.3B, the end-to-end extremely similar@1 metrics improves from 14.9\% to 19.1\% ($\sim$ 28\% relative improvement). With Transformer-based backbone, we can better leverage the largest pretraining dataset Annotations-1.3B, and further improves the metrics to 23.9\% ($\sim$ 60\% relative improvement). Comparing to the current production model, the end-to-end extremely similar@1 metrics improves $\sim$38\% relative.

\subsection{A/B Experiments}
\input{tables/ue_ab_exp}

We take two model variants for online visual shopping A/B experiment - ResNeXt-101 and ViT-B/16 224x pretrained on Annotations-1.3B dataset. The relevance improvement from large-scale pretraining translates to better user engagement as shown in Table ~\ref{tab:ue_ab_exp}. ViT-B/16 224x pretrained on Annotations-1.3B improves both the volume of click-throughs and the number of users who click-through the recommendations by more than 20\%.

Though ViT-B/16 384x scores higher in terms of relevance metrics (Table ~\ref{tab:ue_e2e_relevance_eval}), it is computationally more expensive. For this application, ViT-B/16 224x provides better trade-offs between the retrieval performance and the computational requirements of the model in training and inference.

\subsection{Computational Efficiency}
\label{sec:computational_efficiency}
\input{tables/computational-efficiency}
There are three model architectures that we consider for production deployment: ResNeXt-101, ViT-B/16 224x and ViT-B/16 384x. We benchmark various practical aspects of the computational efficiency on a single GPU shown in Table \ref{tab:computation_efficiency}. The detailed environment setup for running the benchmark is found in the appendix.

For batch operations such as distributed training and batch inference, ResNeXt-101 and ViT-B/16 224x have similar throughput, and thus cost similarly to train the model and compute the embeddings for the visual search corpus. For real-time applications such as visual shopping, we run single image inference. ViT-B/16 224x has the lowest single image inference latency. Overall, we found that ViT-B/16 224x model is the most computationally efficient model to deploy in the production environment.

\subsection{Pretraining Dataset Scale}

\begin{figure}
  \centering
  \includegraphics[width=0.8\linewidth]{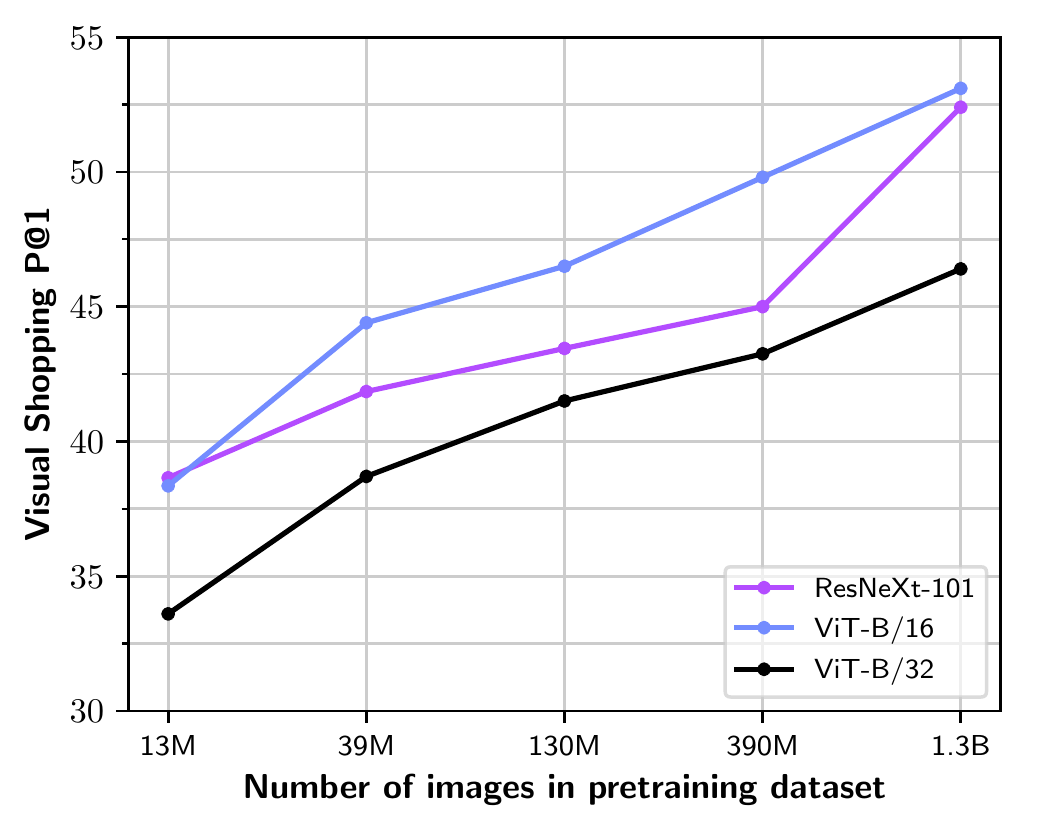}
  \caption{Retrieval performance (Visual Shopping P@1) with respect to the size of the pretraining dataset.}
  \label{fig:dataset-scale}
\end{figure}

The sample count of the pretraining dataset has a significant impact on the Unified Visual Embedding retrieval performance. In Figure~\ref{fig:dataset-scale}, we notice a consistent trend of improvements as the size of the pretraining dataset is increased. When training with 1\% of the Annotations-1.3B dataset, the ResNeXt-101 model architecture has a slight performance advantage (+0.4\%) relative to the ViT-B/16 model architecture. However, as the sample count increases, the performance advantage of the ViT-B/16 model becomes more clear. These findings are consistent with the results in Table~\ref{tab:ue_pretraining_ablation}, where pretraining on ImageNet-1k vs. Annotations 1.3B yields an absolute difference of 12.8\% in VS P@1 performance for the ResNeXt-101 model architecture, whereas the absolute difference in performance of 25.5\% for the ViT-B/32 model architecture is significantly greater. For future work, it would be valuable to study the performance tradeoffs at even larger pretraining dataset sizes, especially with parallel scaling of the Transformer model capacity.

\subsection{Pretraining Label Distribution}
\begin{figure}
  \centering
  \includegraphics[width=0.8\linewidth]{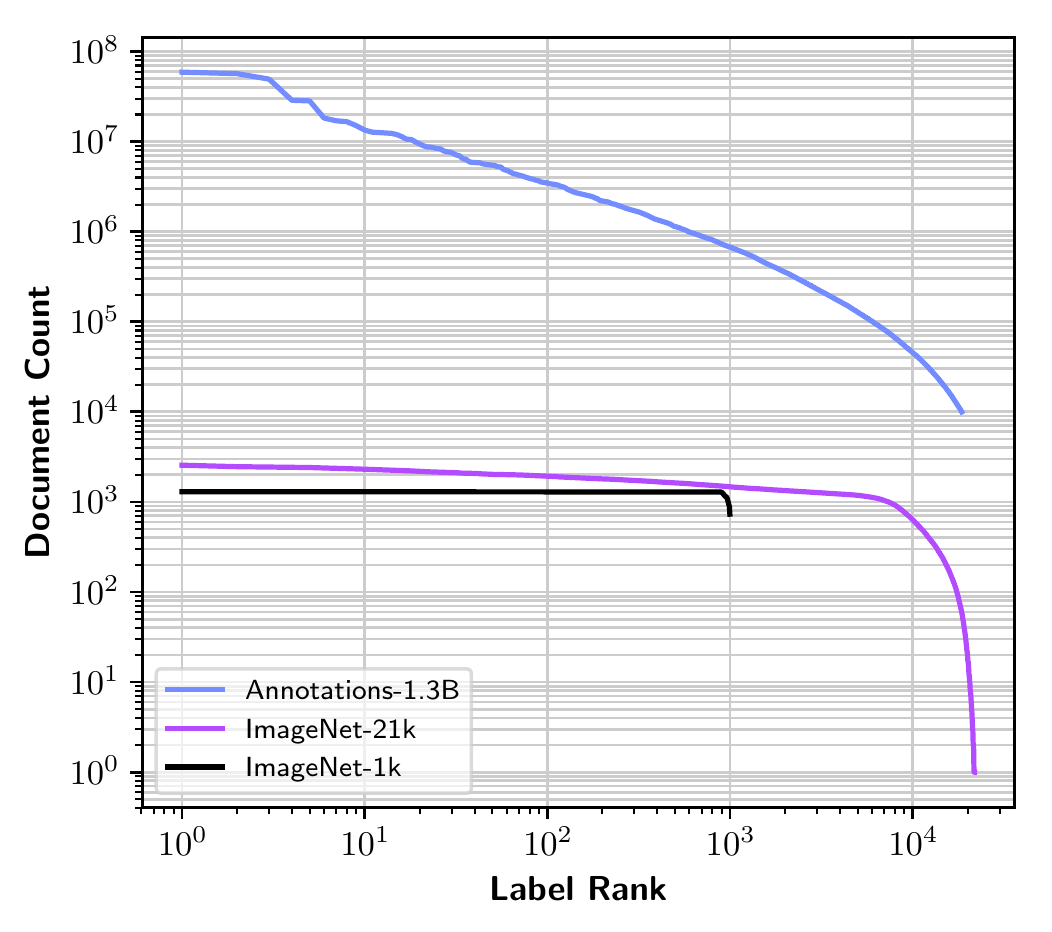}
  \caption{Label distribution of image datasets, including ImageNet-1k, ImageNet-21k, and Annotations-1.3B.}
  \label{fig:pretraining_label_dist}
\end{figure}

Figure~\ref{fig:pretraining_label_dist} highlights the Zipfian~\cite{zipf2013} distribution of Annotations-1.3B and the non-Zipfian distribution of other common pretraining datasets, including ImageNet-1k and ImageNet-21k. To handle the large class imbalance of webly-supervised data, it has been demonstrated in prior work that resampling according to the inverse square root of class frequency yields improvements to the learning of rare classes and can improve the transfer performance of the pretrained representation~\cite{mikolov2013distributed, mahajan2018exploring}. We explored the benefits of this dataset resampling technique for the Annotations-1.3B dataset, finding that resampling the dataset for ViT-B/16 224x pretraining improved the VS P@1 from 41.0\% to 53.1\% as compared to uniform sampling of the dataset.

\subsection{Few-Shot Learning}
\begin{figure}
  \centering
  \includegraphics[width=0.9\linewidth]{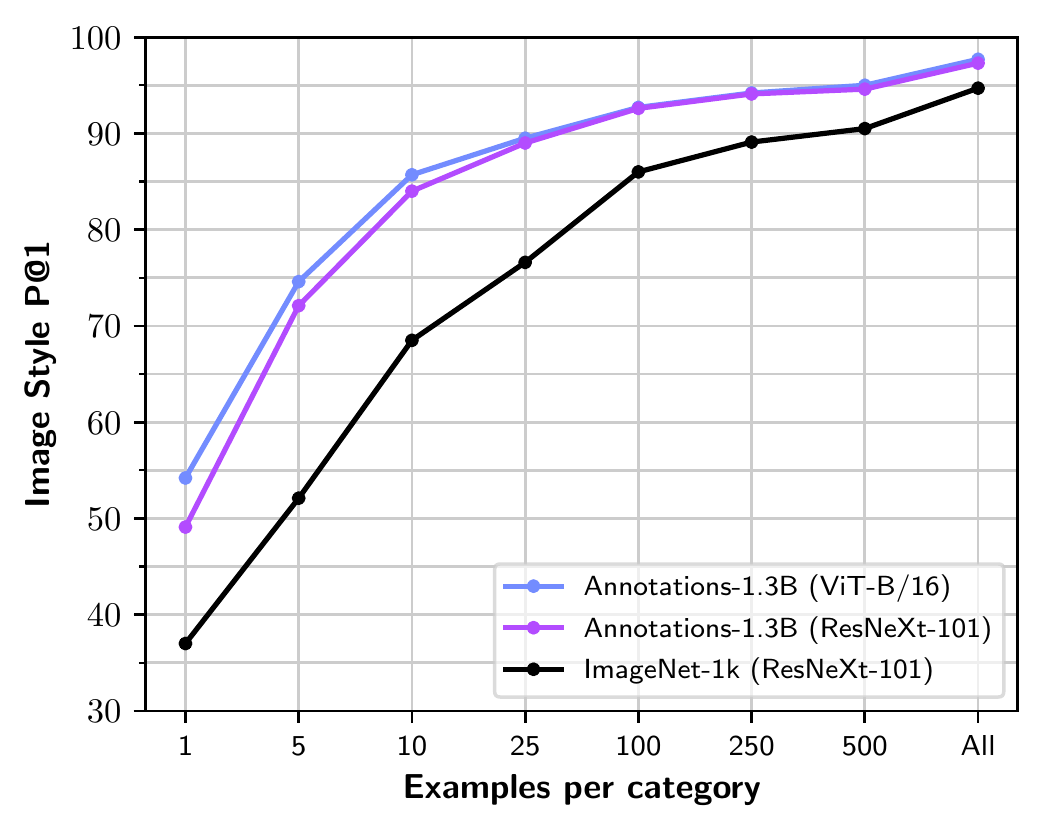}
  \caption{Large-scale pretraining yields better transfer performance in the small data regime for the Image Style task.}
  \label{fig:fewshot}
\end{figure}
Given the significant costs associated with collecting human labels, we study the extent to which pretraining can improve the sample efficiency of fine-tuning as a way to help reduce these costs for future label collection efforts. We consider the Image Style multi-label classification task as an example for our analysis. This dataset consists of 72k images and has an average of 6.2k labels per category. The dataset categories, such as ``Screenshot,'' ``Mosaic,'' and ``Infographic,'' are useful for content understanding at scale. In Figure~\ref{fig:fewshot}, we present the Precision@1 performance for the Image Style task head when varying the number of images per category (1, 5, 10, 25, 100, 250, 500, all) used in fine-tuning. During fine-tuning, the other Unified Visual Embedding dataset sizes remain the same.

As suggested in the context of language modeling \cite{hernandez2021scaling}, large-scale pretraining can serve as an effective multiplier of the size of each fine-tuning dataset. When fine-tuned with only 500 samples per category, the ResNeXt-101 model pretrained on Annotations-1.3B achieves a similar level of performance (94.6\%) as compared to the ResNeXt-101 model pretrained on ImageNet-1k and fine-tuned on the full dataset (94.7\%). Furthermore, when the smaller-size dataset was used in conjunction with the Annotations-1.3B pretrained Vision Transormer backbone, it was able to outperform the full-size dataset used in conjunction with the ImageNet-1k pretrained backbone (95.0\%). In the extreme case of fine-tuning with only 5 samples per category, the large-scale pretrained backbone yields a +20\% absolute improvement.

\subsection{Cross-Domain Generalization}
\input{tables/dataset-scale-transfer}
\begin{figure}
  \centering
  \includegraphics[width=0.9
  \linewidth]{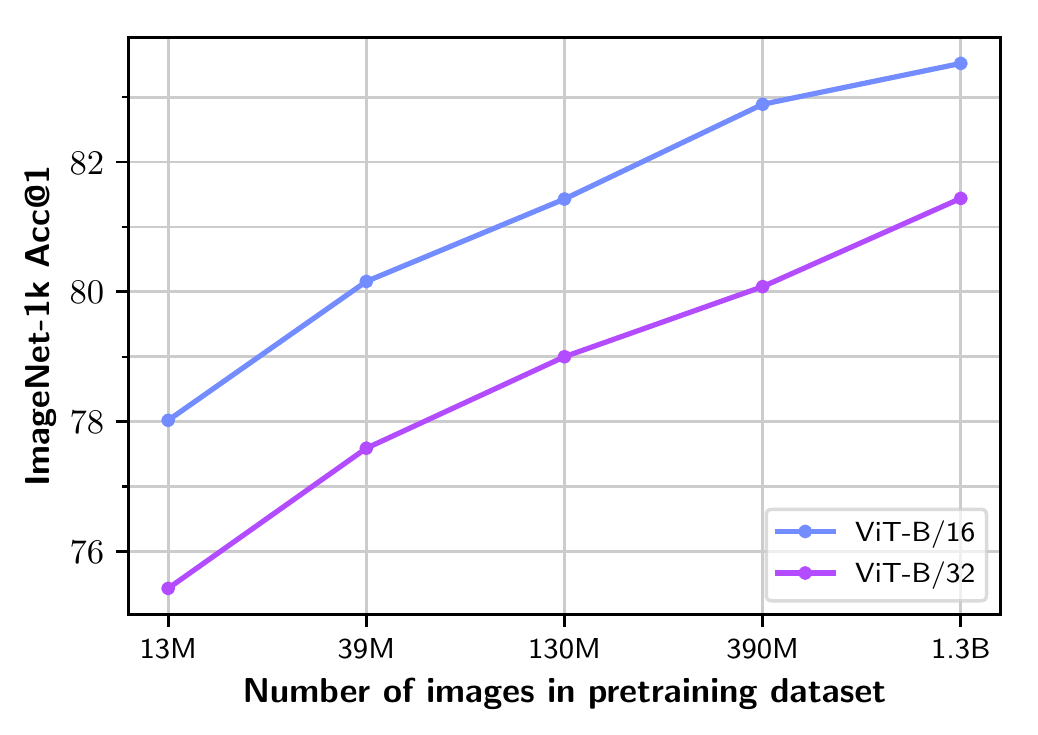}
  \caption{Cross-domain generalization performance with respect to the size of the pretraining dataset.}
  \label{fig:dataset-scale-transfer}
\end{figure}

The ability of the pretrained model to generalize is important as the requirements for the Unified Visual Embedding continue to evolve over time. We study the cross-domain generalization performance of our pretraining approach by analyzing the transfer performance on image classification benchmarks, including ImageNet-1k and ObjectNet~\cite{barbu2019objectnet}, a challenging real-world test set that controls for the biases of rotation, viewpoint, and environment in the dataset construction. In Figure~\ref{fig:dataset-scale-transfer} we see that generalization performance improves as the sample count of the pretraining dataset is increased from 13M to 1.3B and as the spatial resolution is increased in the comparison between ViT-B/32 and ViT-B/16.
Per Table~\ref{tab:dataset_scale_transfer}, without any specific optimization for the ImageNet taxonomy in the dataset construction, Annotations-1.3B achieves competitive transfer performance on the tasks relative to other large-scale image datasets.

\begin{figure}
\begin{center}
\includegraphics[width=0.95\linewidth]{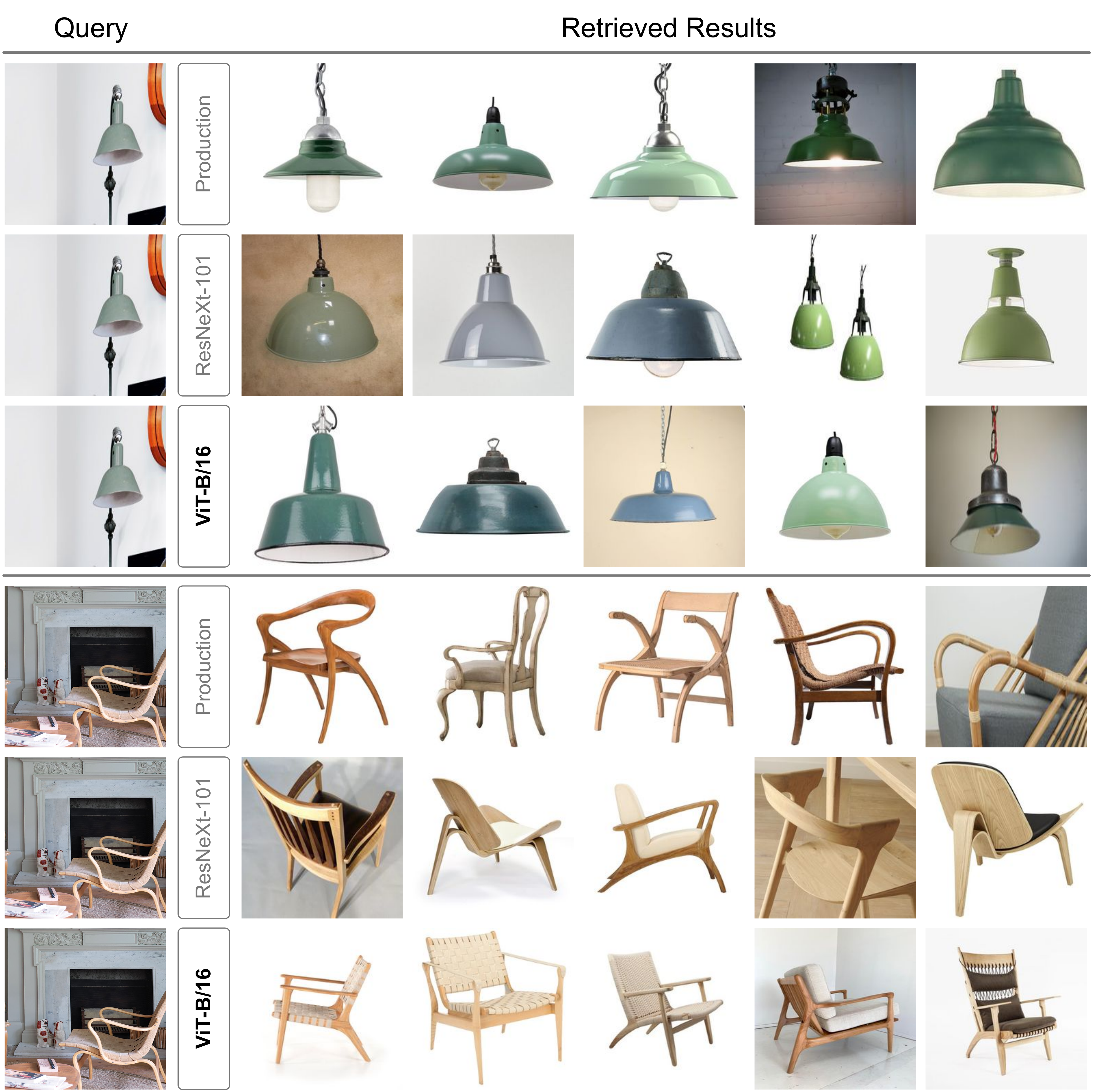}
  \caption{Example of retrieval results using the control (production) model, the ResNeXt-101 Annotations-1.3B model, and the ViT-B/16 Annotations-1.3B model. The ViT model generally matches more similar product results.}
  \label{fig:qual_results}
\end{center}
\end{figure}

\section{Conclusion}

We presented a scalable approach for pretraining with over a billion images in order to improve a production Unified Visual Embedding model. By leveraging heterogeneous sources of textual supervision in a principled fashion, we constructed a large-scale image dataset known as Annotations-1.3B that obtained strong transfer performance and enabled adoption of the state-of-the-art Vision Transformer architecture. The embedding yielded significant improvements to the Visual Shopping system when deployed in production and demonstrated strong advantages across a variety of use cases. This work suggests the promise of further scaling of the Transformer-based pretraining paradigm as a way to systematically improve complex computer vision applications.

\section{Acknowledgements}
The authors would like to thank Eric Tzeng, Raymond Shiau, Kofi Boakye, Vahid Kazemi, and Chuck Rosenberg for valuable discussions regarding the paper, and the anonymous reviewers and ACs for their helpful suggestions.

{\small
\bibliographystyle{ieee_fullname}
\bibliography{egbib}
}

\renewcommand{\thesection}{\Alph{section}}
\setcounter{section}{0}

\section{Appendix}

\subsection{Implementation Details}
\label{subsection:implementation}
Pretraining is implemented using PyTorch on 8 p3dn.24xlarge Amazon EC2 instances with a total of 64 Tesla V100 GPUs, while the fine-tuning uses PyTorch on a single p3dn.24xlarge Amazon EC2 instance with 8 Tesla V100 GPUs. We use DistributedDataParallel for multi-GPU training. We use automatic mixed precision for all of our experiments, and channels-last memory format for the ResNeXt experiments, in order to improve the training throughput. All of our model training runs and performance benchmarks use PyTorch 1.7.1, CUDA 11.0, and cuDNN 8.

Vision Transformer pretraining uses a warmup phase of 10k steps, total batch size of 8192, base learning rate (LR) of 8e-4, and linear decay LR schedule of 2 epochs in length, such that around 2.6B images are processed during the main phase of training. We train using the AdamW~\cite{loshchilov2018decoupled} optimizer with a weight decay value of 0.05. Vision Transformer Unified Visual Embedding fine-tuning uses a warmup phase of 5k steps, base LR of 0.24, and cosine decay LR schedule of 20 epochs in length. We fine-tune using the SGD optimizer with a base LR of 0.24 and weight decay of 1e-4 for the non-sparse parameters. Vision Transformer ImageNet fine-tuning uses a warmup phase of 5k steps, base LR of 0.03, cosine decay LR schedule of 50k steps, SGD optimizer, and zero weight decay.

ResNeXt-101 pretraining uses a warmup phase of 15k steps, total batch size of 12288, base learning rate of 0.03, and step LR schedule of 20 steps and $\gamma = 0.5$. We train using the LARS~\cite{You2017-zc} optimizer with a weight decay value of 1e-4. The hyperparameters of the ResNeXt-101 Unified Visual Embedding fine-tuning are largely the same as ~\cite{Shiau2020-lc}, except the base learning rate is 0.03.

For pretraining, we use the Inception~\cite{szegedy2015going} random crop strategy, whereas for Unified Visual Embedding fine-tuning we apply horizontal mirroring, random crops, and color jitter to the resized images. For ImageNet fine-tuning we directly apply the data augmentation strategy that is specified in the Vision Transformer work.

For ablations on the sample count of the pretraining dataset, we linearly interpolate the training schedule length between the minimum and maximum value, i.e., 100 epochs on the 13M dataset and 2 epochs on the 1.3B dataset.

\end{document}

%% file: tables/ue-pretraining-ablation.tex
\begin{table}
    \centering
    \begin{tabular}{llcccc}
        \toprule
        Model & Pretraining & VS & F & L & C \\
        \midrule
RN-101 & IN-1k & 39.6 & 59.7 & 17.2 & 85.2 \\
RN-101 & IG-940M & 46.7 & 67.6 & 20.2 & 87.9 \\
RN-101 & ANN-1.3B & 52.4 & 70.8 & 22.7 & 88.8 \\
        \midrule
ViT-B/32 & IN-1k & 29.2 & 44.7 & 15.2 & 82.3 \\
ViT-B/32 & ANN-1.3B & 46.4 & 68.9 & 24.9 & 86.5 \\
ViT-B/16 & ANN-1.3B & \textbf{54.7} & \textbf{74.3} & \textbf{26.7} & \textbf{89.7} \\
        \bottomrule
    \end{tabular}
    \caption{Summary of Unified Embedding retrieval performance for different pretraining datasets and model architectures. ``VS'' = Visual Shopping Precision@1, ``F'' = Flashlight Average Precision@20, ``L'' = Lens Average Precision@20, ``C'' = Average Precision@1 of ten image classification tasks. ``RN-101'' = ResNeXt-101, ``IN-1k'' = ImageNet-1k, and ``ANN-1.3B'' = Annotations-1.3B.}
    \label{tab:ue_pretraining_ablation}
\end{table}

%% file: tables/label-generation-ablation.tex
\begin{table}
    \centering
    \begin{tabular}{lcc}
        \toprule
        Method & Label Count & VS  \\
        \midrule
Annotation Clustering & 34k & --- \\
+ Visual Dictionary & 26k & +0.9\% \\
+ L1 Interest Restrict & 18k & +1.1\% \\
        \bottomrule
    \end{tabular}
    \caption{Comparison of retrieval performance and pretraining label count for variants of the label generation procedure. Improvements are relative to variant in previous row.}
    \label{tab:label_generation_ablations}
\end{table}

%% file: tables/ue-resolution-ablation.tex
\begin{table}
    \centering
    \begin{tabular}{llcccc}
        \toprule
        Model & Resolution & VS & F & L & C \\
        \midrule
RN-101 & 224x & 52.4 & 70.8 & 22.7 & 88.8 \\
RN-101 & 384x & 52.3 & 71.4 & 22.5 & 89.0 \\
        \midrule
ViT-B/16 & 224x & 53.1 & 73.7 & 26.3 & 89.2  \\
ViT-B/16 & 384x & \textbf{54.7} & \textbf{74.3} & \textbf{26.7} & \textbf{89.7}  \\
        \bottomrule
    \end{tabular}
    \caption{Summary of retrieval performance for different input image resolutions and model architectures.}
    \label{tab:ue_resolution_ablation}
\end{table}

%% file: tables/ue-e2e-relevance.tex
\begin{table*}
    \centering
    \begin{tabular}{llccc}
        \toprule
        Model & Pretraining & Offline VS P@1 & E2E Extremely Similar@1 & E2E Similar@1 \\
        \midrule
Production~\cite{Shiau2020-lc} & IG-940M & 44.7 & 17.3 & 39.0 \\
        \midrule
ResNeXt-101 & ImageNet-1k & 39.6 & 14.9 & 32.1 \\
ResNeXt-101 & IG-940M & 46.7 & 17.9 & 36.8 \\
ResNeXt-101 & Annotations-1.3B & 52.4 & 19.1 & 38.8 \\
ViT-B/16 224x & Annotations-1.3B & 53.1 & 23.6 & 40.6 \\
ViT-B/16 384x & Annotations-1.3B & \textbf{54.7} & \textbf{23.9} & \textbf{42.3} \\
        \bottomrule
    \end{tabular}
    \caption{Visual Shopping end-to-end retrieval performance using Unified Embedding with different pretraining datasets and model architectures. The end-to-end system retrieval performance is measured by our in-house human evaluators. Two levels of relevance (Extremely Similar and Similar) are shown in the table.}
    \label{tab:ue_e2e_relevance_eval}
\end{table*}

%% file: tables/ue_ab_exp.tex
\begin{table}
    \centering
    \begin{tabular}{l|cc|cc}
        \toprule
        Model & C-vol & CT-vol & C-er & CT-er \\
        \midrule
Production~\cite{Shiau2020-lc} & --- & --- & --- & --- \\
        \midrule
ResNeXt-101 & +10\% & +8\% & +7\% & +7\% \\
ViT-B/16 224x &  \textbf{+22\%} & \textbf{+23\%} & \textbf{+17\%} & \textbf{+22\%} \\
        \bottomrule
    \end{tabular}
    \caption{Visual Shopping A/B experiment using large-scale Annotations-1.3B pretrained Unified Embedding with CNN-based and transformer-based architectures. We report click (C-vol) and click-through volume (CT-vol) relative improvement over our current production system. We also report relative improvement of number of users who click (C-er) or click-through (CT-er) on the product recommendations.}
    \label{tab:ue_ab_exp}
\end{table}

%% file: tables/computational-efficiency.tex
\begin{table}
    \centering
    \begin{tabular}{c|ccc}
        \toprule
        Model & RN-101 & \makecell{ViT 224x} & \makecell{ViT 384x} \\
        \midrule
        \makecell{Train throughput\\ (\#image/s)} & \textbf{265} & 231 & 59 \\
        \midrule
        \makecell{Inference throughput\\ (\#image/s)} & 306 & \textbf{327} & 94 \\
        \midrule
        \makecell{Latency\\ (ms/image)} & 15.6 & \textbf{8.7} & 13.3 \\
        \bottomrule
    \end{tabular}
    \caption{Comparison of computation efficiency for different model architectures. Train and inference throughput are measured with maximum batch size for a single GPU. Latency refers to single-image batch inference. ``ViT'' = ViT-B/16.}
    \label{tab:computation_efficiency}
\end{table}

%% file: tables/dataset-scale-transfer.tex
\begin{table}
    \centering
    \begin{tabular}{llccccc}
        \toprule
        Model & Pretraining & ImageNet-1k & ObjectNet \\
        \midrule
ViT-B/32 & ANN-1.3B & 81.4 & 48.4 \\
ViT-B/32 & JFT-300M & 80.7 & --- \\
        \midrule
ViT-B/16 & ANN-1.3B & 83.6 & 50.7 \\
ViT-B/16& JFT-300M & 84.1 & --- \\
        \midrule
ResNet-50 & JFT-300M & 77.5 & 42.5 \\
ResNet-101 & JFT-300M & 80.6 & 49.1 \\
ResNeXt-101 & IG-940M & 82.2 & --- \\
        \bottomrule
    \end{tabular}
    \caption{
Cross-domain generalization performance (ImageNet-1k and ObjectNet top-1 accuracy) for different pretraining approaches and model architectures \cite{mahajan2018exploring, kolesnikov2019big, dosovitskiy_arxiv2020} .
}
    \label{tab:dataset_scale_transfer}
\end{table}